\documentclass[11pt,a4paper]{article}
\usepackage[final]{emnlp2021}
\usepackage{times}
\usepackage{latexsym}

\usepackage[utf8]{inputenc} 
\usepackage[T1]{fontenc}    
\usepackage{hyperref}       
\usepackage{url}            
\usepackage{booktabs}       
\usepackage{amsfonts}       
\usepackage{nicefrac}       
\usepackage{microtype}      
\usepackage{amsmath,amssymb}
\usepackage{algpseudocode}
\usepackage{multirow,booktabs, hhline}
\usepackage[ruled,noend]{algorithm2e}
\usepackage{amsmath, bm}
\usepackage{wrapfig}
\usepackage{lipsum}
\usepackage{caption}
\usepackage{footnote}
\usepackage{color}
\usepackage[bottom]{footmisc}
\makesavenoteenv{tabular}
\usepackage[toc,page]{appendix}

\usepackage{graphicx}
\usepackage{amsmath}
\usepackage{amsthm}
\usepackage{booktabs}
\usepackage{xspace}
\usepackage{CJKutf8}
\usepackage{microtype}
\usepackage{subcaption}
\usepackage{blindtext}

\newcommand{\abbr}{SSR\xspace}
\newcommand{\paratitle}[1]{\noindent\textbf{#1}}

\usepackage{microtype}
\usepackage{subcaption}

\title{Improving Sequence-to-Sequence Pre-training via \\ Sequence Span Rewriting}

\author{
Wangchunshu Zhou$^{1}$\thanks{\ \ This work was done during the first author's internship at Microsoft Research Asia.} ~~~ Tao Ge$^{2}$ ~~~ Canwen Xu$^{3}$ ~~~ Ke Xu$^{4}$ ~~~ Furu Wei$^2$\\
$^1$Stanford University ~~ $^2$Microsoft Research Asia \\
$^3$University of California, San Diego ~~ $^4$Beihang University\\
{\tt wcszhou@stanford.edu.cn, cxu@ucsd.edu kexu@nlsde.buaa.edu.cn}\\
{\tt \{tage, fuwei\}@microsoft.com}}

\date{}

\begin{document}
\maketitle

\begin{abstract}
In this paper, we propose \textbf{S}equence \textbf{S}pan \textbf{R}ewriting (SSR), a self-supervised task for sequence-to-sequence (Seq2Seq) pre-training. SSR learns to refine the machine-generated imperfect text spans into ground truth text. SSR provides more fine-grained and informative supervision in addition to the original text-infilling objective. Compared to the prevalent text infilling objectives for Seq2Seq pre-training, SSR is naturally more consistent with many downstream generation tasks that require sentence rewriting (e.g., text summarization, question generation, grammatical error correction, and paraphrase generation). We conduct extensive experiments by using SSR to improve the typical Seq2Seq pre-trained model T5 in a continual pre-training setting and show substantial improvements over T5 on various natural language generation tasks.\footnote{Code for pre-training SSR is available at \url{https://github.com/MichaelZhouwang/Sequence_Span_Rewriting}.} 

\end{abstract}
\section{Introduction}

\begin{figure*}[t]
    \centering
    \includegraphics[width=16cm]{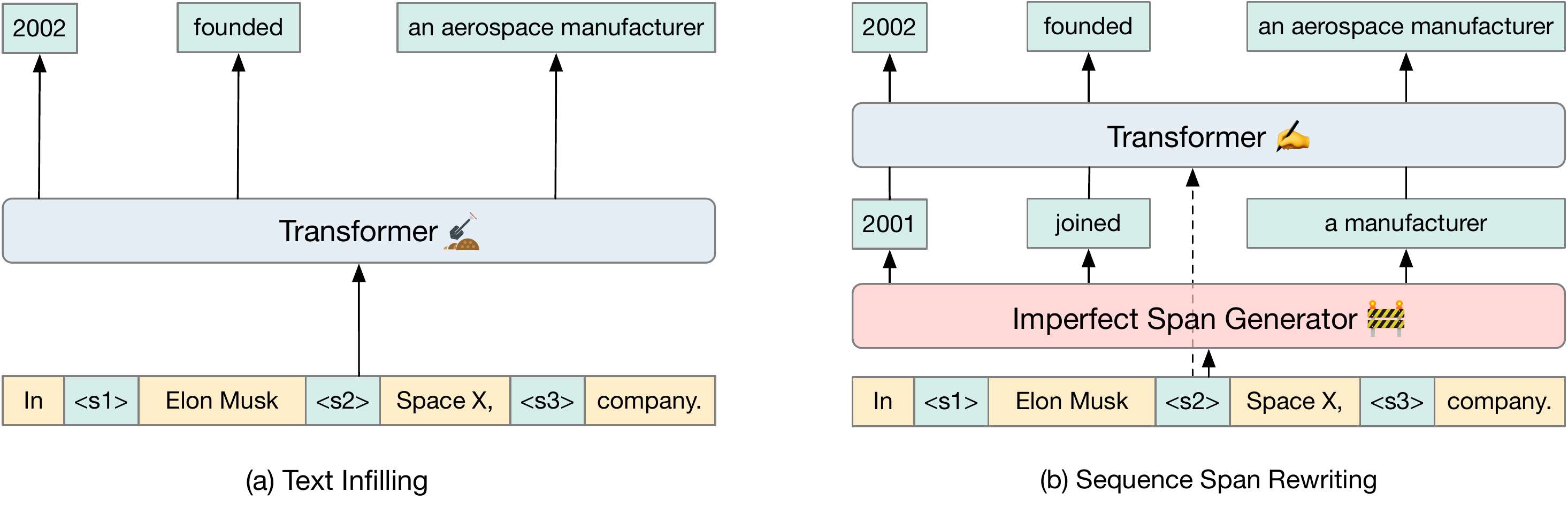}
    \caption{The comparison of \textbf{(a)} Text Infilling and \textbf{(b)} Sequence Span Rewriting. Instead of learning to directly fill the blanks, Sequence Span Rewriting first exploits an imperfect span generator to generate imperfect spans within the text and then feeds the filled text to the model to learn how to rewrite it into the ground truth.}
    \label{fig:mainexample}
\end{figure*}

Text infilling (e.g., masked language modeling) has become a prevalent learning objective for pre-trained language models (PTLMs)~\citep{peters2018deep,radford2018improving,devlin2018bert,yang2019xlnet,liu2019roberta,lan2019albert,lewis2019bart,raffel2019exploring}. It provides self-supervision by masking out tokens or spans in text, and trains a model to infill the masked content based on the contexts, accordingly guiding the model for representation learning, as Figure \ref{fig:mainexample}(a) shows.

In this paper, we propose to extend
the conventional text infilling to a novel sequence-to-sequence (Seq2Seq) pre-training objective, namely \textbf{S}equence \textbf{S}pan \textbf{R}ewriting (SSR). We train a model to rewrite machine-generated imperfect text spans into the ground truth text, as illustrated in Figure \ref{fig:mainexample}(b).
SSR has two advantages over text infilling: \textbf{(1) SSR provides better supervision signals,} as SSR trains the model with diverse and fine-grained rewriting patterns beyond filling the blanks; \textbf{(2) SSR bridges the gap between pre-training and fine-tuning,} because many downstream Seq2Seq tasks like summarization and paraphrase generation are naturally sequence span rewriting tasks where a source sentence is mapped to the target sentence following specific rewriting patterns.

The key element in implementing SSR is how to generate imperfect text spans that are both diverse and informative.
Inspired by ELECTRA~\cite{clark2020electra}, we use a powerful pre-trained text infilling model -- T5-large~\cite{raffel2019exploring} -- as the imperfect span generator. Compared with random or rule-based noising approaches, the T5-based imperfect span generator can derive various informative text spans that benefit the model to learn meaningful and diverse rewriting patterns including paraphrasing and enhancing the fluency and contextual consistency through correcting grammatical, commonsense and factual errors, to improve a text sequence. These rewriting patterns resemble the goal of various NLG tasks and thus strengthen the ability of pre-trained model for downstream applications.



In our experiments, we apply SSR to the typical Seq2Seq pre-trained model -- T5~\cite{raffel2019exploring} in a continual learning fashion.
We show SSR outperforms both the original pre-trained T5 models and their continual training counterparts with the conventional text infilling objective on various Seq2Seq tasks, including text summarization, question generation, and grammatical error correction, with a small number of optimization steps with moderate amount of machine-generated data, which confirms the potential of SSR to serve as a plug-and-play method to improve various existing pre-trained Seq2Seq models.
Notably, we find SSR especially useful for pre-training smaller Seq2Seq models, with the help of a powerful imperfect span generator. 
This observation sheds light on a new approach for knowledge transfer from large models to smaller ones.

\section{Related Work}

\paragraph{Pre-training in NLP}

BERT~\citep{devlin2018bert} introduced the masked language modeling objective by masking out certain tokens in a text and predicting them based on their left and right side contexts.  Recent work has shown that BERT's performance can be further improved by training for longer~\citep{liu2019roberta}, by tying parameters across layers~\citep{lan2019albert}, and by replacing a consecutive span of tokens with the mask token for MLM training~\citep{DBLP:journals/tacl/JoshiCLWZL20}. Our approach is also related to ELECTRA~\cite{clark2020electra}, which uses a pre-trained masked language model to generate fake tokens and train a discriminator to detect them. The key difference is that our approach focuses on span-level texts and trains the model to correct the mistakes instead of simply detecting them, which includes more diverse and informative signals and enables the model to perform text generation tasks in a Seq2Seq fashion. 

To enable mask language models for natural language generation tasks,~\citet{song2019mass} used a decoder to generate the masked tokens autoregressively. UniLM~\citep{dong2019unified} multitasks MLM and language modeling objectives. More recently, BART~\citep{lewis2019bart} and T5~\citep{raffel2019exploring} pre-train Seq2Seq models with the text span infilling objective, which removes text spans in the input texts and train the models to recover the original texts in an auto-regressive fashion.


More recently, CALM~\citep{DBLP:conf/iclr/ZhouLSL021} introduces concept-to-sentence generation and concept order recovery as two self-supervised objectives that encourage Seq2Seq PTLMs to acquire generative commonsense reasoning ability. MARGE~\citep{lewis2020pre} pre-trains a Seq2Seq model with an unsupervised multi-lingual cross-document
paraphrasing objective. Their approach is related to our text rewriting objective. However, MARGE requires multi-lingual paraphrase documents and needs to train a separate retrieval model while our method can simply used an off-the-shelf model pre-trained with text infilling to generate training data. Also, MARGE is pre-trained to generate a paraphrase-like document in another language, thus mainly helpful for translation tasks and multi-lingual tasks. In contrast, SSR focus on monolingual text rewriting and improve general text generation tasks.

SSR departs significantly from the aforementioned methods for Seq2Seq pre-training as it employ machine-generated noises instead of rule-based ones, thus introducing more diverse training signals. Also, SSR receives complete inputs without artificial masks during pre-training relying solely on monolingual corpus.

\paragraph{Model Acceleration for PTLMs}
Recently, many attempts have been made to speed up a large pre-trained language model (PTLM). To name a few, \citet{shen2020q} quantized BERT to 2-bit using Hessian information; \citet{michel2019sixteen} pruned unnecessary attention heads in the transformer layers to reduce the parameters of a BERT model. DistilBERT~\citep{sanh2019distilbert} and uses knowledge distillation~\cite{hinton2015distilling,romero2014fitnets} to compress BERT. More recently,~\citep{DBLP:journals/corr/abs-2106-04570} proposed Meta Distillation to improve the performance of knowledge distillation for compression BERT. In addition,~\citet{xu2020bert} introduced progressive module replacing to train more compact BERT models by encouraging the student model to behave similarly with the teacher model.
In addition,~\citet{zhou2020bert,schwartz2020right} proposed to accelerate the inference stage of pre-trained models via input-adaptive inference.
However, to the best of our knowledge, few studies have been done for accelerating large sequence-to-sequence PTLMs.
Our approach can also be used for model compression by using a large pre-trained model as the imperfect span generator. In this way, SSR also exploits the knowledge of a larger model to improve the training of a compact model.
\section{Methodology}

The key idea of SSR is to train a Seq2Seq model to rewrite machine-generated text spans that may contain a variety of noise such as paraphrase, grammatical and factual errors, into ground truth that are correct and appropriate in the context. As illustrated by Figure \ref{fig:mainexample}(b), SSR involves three steps: (1) masking out parts of the text; (2) generating imperfect text to fill in the masked spans; (3) training the Seq2Seq model to rewrite the imperfect spans to the ground truth. We will introduce the technical details of SSR in Section \ref{subsec:ssr} and an advanced training strategy for SSR in Section \ref{subsec:curriculum}.

\subsection{Sequence Span Rewriting} \label{subsec:ssr}

\paragraph{Text Span Masking} To generate training data of sequence span rewriting in a self-supervised fashion, we first randomly sample a number of text spans and mask them. Specifically, the spans are masked with special mask tokens by order (e.g., \texttt{<s$_1$>}, \texttt{<s$_2$>} and \texttt{<s$_3$>}) in Figure \ref{fig:mainexample}(b) as in T5, with span lengths drawn from a Poisson distribution ($\lambda$ = 3). The number of spans is controlled so that approximately 30\% of all tokens are masked. Specially, 0-length spans correspond to an insertion of a mask token.

For example, as shown in Figure \ref{fig:mainexample}, given a sentence ``In 2002, Elon Musk founded SpaceX, an aerospace manufacturer company.'', we randomly sample three text spans (two of them are of length 1). The masked sentence becomes ``In \texttt{<s$_1$>}, Elon Musk \texttt{<s$_2$>} SpaceX, \texttt{<s$_3$>} company.''

\paragraph{Imperfect Span Generation} With masked spans, we can generate imperfect text to fill in the spans. Specifically, we feed the masked input into the imperfect span generator to generate predictions in an auto-regressive fashion. To improve the diversity of generation, we use nucleus sampling~\citep{Holtzman2020The} that truncates the unreliable tail of the probability distribution and samples from the dynamic nucleus of tokens containing the vast majority of the probability mass. For instance, given the previous masked input sentence, a T5-large model generates ``2001'', ``joined'', and ``a manufacturer'' as imperfect spans.

\paragraph{Span Rewriting} After we obtain imperfect spans within the text, we pre-train the Seq2Seq model to rewrite imperfect text spans into the ground truth. Specifically, we use special tokens \texttt{<s$_i$>} and \texttt{</s$_i$>} to denote the starting and ending of $i$-th text span to be rewritten in the source sequence, which gives  ``In \texttt{<s$_1$>} 2001 \texttt{</s$_1$>}, Elon Musk \texttt{<s$_2$>} joined \texttt{</s$_2$>} SpaceX, \texttt{<s$_3$>} a manufacturer \texttt{</s$_3$>} company.'' as the input for SSR pre-training.
Similarly, we use \texttt{<s$_i$>} to separate different text spans in the target sequence, which gives ``\texttt{<s$_1$>} 2002 \texttt{<s$_2$>} founded \texttt{<s$_3$>}  an aerospace manufacturer'' as the target sequence. 
We train the model to generate target text spans from left to right auto-regressively by maximum likelihood estimation. 

We can see that the SSR objective involves using a pre-trained model to generate imperfect spans, which will lead to increased computational cost. In practice, we suggest starting SSR pre-training based on checkpoints of existing Seq2Seq pre-trained models. In this way, we only need to generate a few amount of imperfect spans and continually pre-train the models for a few steps. In this perspective, SSR can be viewed as a general approach that be used to improve various Seq2Seq pre-trained models before fine-tuning them on downstream text generation tasks.

For fine-tuning SSR, we simply denote the entire input sequence with the same span identifier (e.g., \texttt{<s$_1$>}) used during SSR pre-training. Therefore, the model would learn to rewrite the entire input sequence, alleviating the gap caused by the \texttt{<mask>} token during text infilling pre-training. For example, for grammatical error correction, the input is formatted as ``\texttt{<s$_1$>} I go to school yesterday. \texttt{</s$_1$>}'' and the output is ``\texttt{<s$_1$>} I went to school yesterday.'', which exactly corresponds to the pre-training format of SSR. 

In addition, for some constrained text generation tasks~\citep{DBLP:conf/emnlp/LinZSZBCR20} and controlled text generation~\citep{DBLP:conf/icml/HuYLSX17} tasks, we can specify which part of input text to be rewritten with span identifiers. This enables more flexible text generation with Seq2Seq pre-trained models. Taking text attribute transfer as an example, an input example would looks like ``Great food \texttt{<s$_1$>} but very rude \texttt{</s$_1$>}  waiters.'' and the corresponding target sequence is ``\texttt{<s$_1$>} and very friendly''. The inductive bias of span rewriting learned by SSR pre-training naturally benefit these kind of NLG applications.

\subsection{Curriculum SSR} \label{subsec:curriculum}
As mentioned above, we apply SSR as a continual training objective for  pre-trained Seq2Seq models that were originally trained with the text infilling objective. However, continually training a pre-trained Seq2Seq model with a different objective may result in drastic adaption of its parameters. To make this transition smoother and reduce the difficulty of optimization, we propose to schedule the SSR training examples with curriculum learning~\citep{bengio2009curriculum} according to their difficulties. Specifically, we measure the difficulty of rewriting a certain imperfect text span with both the length of the imperfect span and the uncertainty (i.e., perplexity) of the imperfect span generator when generating this span. 

Intuitively, a short imperfect span generally includes some simple word substitution (e.g., big $\rightarrow$ large) or grammatical error (e.g., is $\rightarrow$ was) while a longer imperfect span may require more complicated paraphrasing (e.g., what is happening $\rightarrow$ what's up). Also, an imperfect span with larger perplexity suggests the span may be of lower quality or more uncommon, thus more difficult to be rewritten into ground truth. Therefore, we consider longer imperfect spans and spans with a higher perplexity under the imperfect span generator to be more difficult. We split the SSR training examples into $k$ ($k=5$ in our experiments) groups according to the sum of per-token loss of the imperfect span generator when it generates an SSR training example. We then start pre-training the model with the easiest group of SSR training examples and then gradually switch to more difficult groups during pre-training. Intuitively, this will make the transition from the original text infilling objective to the sequence span rewriting objective more smooth. 
\section{Experiments}
 
\subsection{Experimental Settings}
 
\abbr is implemented as a text-to-text transformer model with a bidirectional encoder and a left-to-right auto-regressive decoder. For pre-training, we minimize the negative log-likelihood of the original ground truth text spans. We describe details of the architecture, pre-training, and fine-tuning of \abbr in this section. 

\paragraph{Architecture}

We use the same architecture as T5~\citep{raffel2019exploring} which is roughly equivalent to the original Transformer proposed by~\citet{vaswani2017attention}, with the exception of removing the Layer Norm bias, placing the layer normalization outside the residual path, and using a different relative position embedding scheme.

Following the design choice of T5~\citep{raffel2019exploring}, we train three sizes of \abbr:
\begin{itemize}
    \item \abbr-small: 60M parameters, 6 Transformer layers, 8 attention heads, 512 hidden size
    \item \abbr-base: 220M parameters, 12 Transformer layers, 12 attention heads, 768 hidden size
    \item \abbr-large: 770M parameters, 24 Transformer layers, 16 attention heads, 1024 hidden size
\end{itemize}


\paragraph{Pre-training Details}
As we propose SSR to serve as a general plug-and-play approach to improve existing Seq2Seq pre-trained models without intensive computation like pre-training from scratch, we initialize each size of \abbr model with the corresponding pre-trained T5 model of the same size, and continually pre-train the models with the SSR objective. 

\begin{table*}[!tbp]
\centering
    \begin{minipage}{\textwidth}
	\centering
	\resizebox{\textwidth}{!}{
		\begin{tabular}{lclllllll}
			\toprule
\multirow{2}{*}{\textbf{Model}} & \multirow{2}{*}{\textbf{Architecture}} & \multicolumn{3}{c}{\textbf{CNN/DM}} &\multicolumn{3}{c}{\textbf{XSum}}\\
            \cmidrule(lr){3-5} \cmidrule(lr){6-8}
			  & & RG-1 & RG-2 & RG-L  & RG-1 & RG-2 & RG-L \\
			\midrule
			\multicolumn{8}{c}{\textit{Performance of models without pre-training}} \\
			\midrule
			Lead-3 & - & 40.42 & 17.62 & 36.67 & 16.30 & 1.60 & 11.95  \\
			PTGEN~\citep{see2017get} & -  & 36.44 & 15.66 & 33.42 & 29.70 & 9.21 & 23.24 \\
			\midrule
			\multicolumn{8}{c}{\textit{Performance of state-of-the-art models based on pre-trained models of comparable size}} \\
			\midrule
			MASS~\citep{song2019mass} & L=6, H=1024 &  42.12 & 19.50 &  39.01 & 39.75 & 17.24 & 31.95 \\
			BERTSumAbs~\citep{liu2019text} & L=12, H=768 & 41.72 & 19.39 & 38.76 & 38.76 & 16.33 & 31.15  \\
			UniLMv2~\citep{bao2020unilmv2} & L=12, H=768 & 43.16 & 20.42 & 40.14 & 44.00 & 21.11 & \bf 36.08 \\
			\midrule
			\multicolumn{8}{c}{\textit{Performance of comparable models based on T5-base}} \\
			\midrule
			T5-base~\citep{raffel2019exploring} & L=12, H=768 & 42.25 & 20.22 & 39.45 & 43.12 & 20.84 & 34.98  \\
			T5-base-cont & L=12, H=768 & 42.49 & 20.33 & 39.65 & 43.32 & 20.94 & 35.21  \\ 
			DistilT5-base & L=12, H=768 & 42.37 & 20.25 & 39.53 & 43.25 & 20.89 & 35.14  \\ 
			DenoiseT5-base & L=12, H=768 & 42.22 & 20.18 & 39.41 & 43.14 & 20.82 & 35.03  \\ 
			\abbr-base  & L=12, H=768 & \bf 43.53$^{*}$ & \bf 20.79$^{*}$ & \bf 40.44$^{*}$ & \bf 44.05 & \bf 21.19 & 35.88 \\
			\bottomrule
	\end{tabular}
	}
	\caption{Abstractive summarization results. We also present the transformer architecture for the methods using pre-trained models. For example, L=12, H=768 means both the encoder and decoder are built with 12 transformer layers with a hidden size of 768. $^{*}$The asterisk denotes statistically significant improvement with p-value < 0.05 upon all compared models.}
	\label{tab:result1}
	\end{minipage}
\end{table*}

For imperfect span generation, we use the off-the-shelf T5-large model
with nucleus sampling ($p=0.9$) to sample generated text spans. For SSR learning, we sample 4GB of text from Wikipedia corpus, BookCorpus~\citep{zhu2015aligning}, and RealNews~\citep{zellers2019defending}, which are commonly used for pre-training language models. Our implementation is based on Hugging Face Transformers~\cite{wolf2020transformers}. We use text sequences with a maximum length of 256 tokens to sample masked text spans and generate imperfect text spans. We then continually pre-train different variants of \abbr for 100k updates\footnote{We empirically find 100k updates to be enough since the models' performance on downstream tasks begin to saturate.}, with a maximum sequence length of 256, a batch size of 512, and a learning rate of 5e-5 with a linear warm-up for the first 8,000 updates.  


It is noteworthy that although SSR requires using a pre-trained Seq2Seq model for imperfect span generation, the computation cost of using SSR to improve a Seq2Seq pre-trained model is still considerably smaller than the pre-training cost. This is because SSR requires much smaller training corpus and optimization steps when employed in a continual pre-training setting. This also reduces recent concerns~\citep{strubell2019energy,parrot} about the carbon footprint and energy consumption in LM pre-training.

\subsection{Tasks and Datasets}


\paratitle{Abstractive Summarization} aims to rewrite a long document into a short summary. 
To provide a comparison with the recent work in pre-trained models for this task, we present results on two widely used summarization datasets: CNN/DailyMail~\citep{hermann2015teaching} and XSum~\citep{narayan2018don}, and report evaluation results in terms of ROUGE-1, ROUGE-2 and ROUGE-L~\citep{lin-2004-rouge}. 

\paratitle{Question Generation} is to generate valid and fluent questions according to a given passage and target answers. It can be considered as rewriting a target answer and its surrounding context into a question form. Following previous work~\citep{dong2019unified}, we concatenate the passage and an answer as the input of the model to learn to generate the corresponding question in the fine-tuning stage. We use SQUAD~\citep{rajpurkar2016squad} dataset to train and test question generation following the data split in~\citep{du-cardie-2018-harvesting}. We report evaluation results in terms of BLEU~\citep{papineni-etal-2002-bleu}, METEOR~\citep{banerjee-lavie-2005-meteor}, and CIDEr~\citep{vedantam2015cider}.

\paratitle{Grammatical Error Correction} is a task that rewrites a potentially erroneous input sentence into a fluent sentence that is grammatical error free without changing the original meaning of the input sentence. Following the recent work~\cite{grundkiewicz2019neural,kiyono2019empirical,zhou2019improving} in GEC, we use the public Lang-8~\cite{mizumoto2011mining}, NUCLE~\cite{dahlmeier2013building}, FCE~\cite{yannakoudakis2011new} and W\&I+LOCNESS datasets~\cite{bryant2019bea,granger1998computer} for fine-tuning without using any synthetic GEC data, and then evaluate Max-Match (M$^2$) precision, recall, and F$_{0.5}$ score on the CoNLL-2014~\citep{ng2014conll} test set.

\begin{table*}[!tbp]
\centering
    \begin{minipage}{\textwidth}
	\centering
	\resizebox{\textwidth}{!}{
		\begin{tabular}{lclllllll}
			\toprule
\multirow{2}{*}{\textbf{Model}} & \multirow{2}{*}{\textbf{Architecture}} & \multicolumn{3}{c}{\textbf{Question Generation}} &\multicolumn{3}{c}{\textbf{GEC}}\\
            \cmidrule(lr){3-5} \cmidrule(lr){6-8}
			  & & BLEU-4 & METEOR & CIDEr  & P & R & F$_{0.5}$ \\
			\midrule
			\multicolumn{8}{c}{\textit{Performance of baseline models without pre-training}} \\
			\midrule
			\citet{zhang2019addressing} & - & 18.37 & 22.65 & 46.68 & - & - & -  \\
			Xfmr-big~\citep{chen2020improving} & L=12, H=1024  & - & - & - & 64.9 & 26.6 & 50.4 \\
			Xfmr-big + Synthetic Data~\citep{zhou-etal-2020-pseudo} & L=12, H=1024  & - & - & - & 69.1 & 33.7 & 57.1 \\
			\midrule
			\multicolumn{8}{c}{\textit{Performance of state-of-the-art models based on pre-trained models of comparable size}} \\
			\midrule
			UniLMv2~\citep{bao2020unilmv2} & L=12, H=768 & \bf 24.43 & 26.34 & 51.97 & - & - & - \\
			\midrule
			\multicolumn{8}{c}{\textit{Performance of comparable models based on T5-base}} \\
			\midrule
			T5-base~\citep{raffel2019exploring} & L=12, H=768 & 23.74 & 25.95 & 51.61 & 68.6 & 33.5 & 56.7  \\
			T5-base-cont & L=12, H=768 & 23.93 & 26.11 & 51.78 & 69.6 & 33.6 & 57.3  \\ 
			DistilT5-base & L=12, H=768 & 23.86 & 25.93 & 51.64 & 69.3 & 33.1 & 56.9  \\ 
			DenoiseT5-base & L=12, H=768 & 23.70 & 25.91 & 51.58 & 69.5 & 33.4 & 57.1  \\ 
			\abbr-base  & L=12, H=768 & 24.35	& \bf 26.51$^{*}$	& \bf 52.11$^{*}$ & \bf 70.5$^{*}$ & \bf 34.9$^{*}$ & \bf 58.7$^{*}$ \\
			\bottomrule
	\end{tabular}
	}
	\caption{Question generation and GEC results. We also present the transformer architecture for the methods using transformer models. For example, L=12, H=768 means both the encoder and decoder are built with 12 transformer layers with a hidden size of 768. $^{*}$The asterisk denotes statistically significant improvement with p-value < 0.05 upon all compared models.}
	\label{tab:result2}
	\end{minipage}
\end{table*}

\subsection{Compared Models}

We compare \abbr with the following models:
\begin{itemize}
    \item \textbf{T5}: the original pre-trained text-to-text transformer based on the text infilling objective.
    \item \textbf{T5-cont}: text-to-text transformer initialized by T5 and continually pre-trained with the original text infilling objective with additional training steps. The total number of additional training steps is equal to that of \abbr.
    \item \textbf{DistilT5}: the variant that continually pre-trains T5 by text infilling with sequence-level knowledge distillation~\citep{kim2016sequence}. This is implemented by using the imperfect text spans generated by T5-large as target outputs for text infilling. DistilT5-small and DistilT5-base are similar to conventional sequence-level knowledge distillation while DistilT5-large can be viewed as continually pre-trained with self-distillation.
    \item \textbf{DenoiseT5}: the variant that injects rule-based noises into plain text and continually pre-train a T5 model to output the original text. The rule-based noises include token shuffling, deletion, and replacement. We adopt the same noise strategy as described in~\citet{wang-etal-2019-denoising}.
\end{itemize}

For reference, we also compare against two state-of-the-art base-sized pre-trained models for NLG including MASS~\citep{song2019mass} and UniLMv2~\citep{bao2020unilmv2}.

\subsection{Experimental Results}

We first present experimental results of \abbr-base and comparable baselines on different datasets. Then we show additional results of \abbr-small and \abbr-large for further analysis.

\paragraph{Summarization Results} 
According to Table \ref{tab:result1}, it is observed that \abbr-base substantially improves the original T5-base model and its continual training variants on both CNN/DM and XSum datasets, and achieves state-of-the-art results for the models of the same size in the abstractive summarization benchmarks. It is notable that our models are only continually pre-trained on a relatively small dataset for only a few number of updates. This confirms the potential of our approach as a general ``plug-and-play'' approach for improving various kinds of sequence-to-sequence pre-trained models.
In contrast, using T5-large as a teacher model fails to improve the training of a T5-base student with sequence-level knowledge distillation. This shows SSR can better exploit the capability of a large Seq2Seq pre-trained model to improve a smaller one, indicating its potential to serve as a model compression technique for Seq2Seq pre-trained models.


\begin{table}[!tbp]
\centering
    \begin{minipage}{\linewidth}
	\centering
	\resizebox{0.95\linewidth}{!}{
		\begin{tabular}{llll}
			\toprule
\multirow{2}{*}{\textbf{Model}} & \multicolumn{3}{c}{\textbf{CNN/DM}} \\
			  & RG-1 & RG-2 & RG-L \\
			\midrule
			T5-large & 43.09 & 20.68 & 40.15 \\
			T5-large-cont & 43.14 & 20.71 & 40.21 \\
			DistilT5-large & 43.05 & 20.63 & 40.07 \\
            \abbr-large & \bf 43.65$^{*}$ & \bf 20.98$^{*}$ & \bf 40.69$^{*}$ \\
			\bottomrule
	\end{tabular}
	}
	\caption{Abstractive summarization results on CNN/DailyMail for SSR-large and corresponding T5 models of the same size. $^{*}$The asterisk denotes statistically significant improvement with p-value < 0.05 upon all compared models. }
	\label{tab:result3}
	\end{minipage}
\end{table}

\paragraph{Question Generation and GEC Results} 

Similar results are observed for question generation and GEC tasks, as shown in Table \ref{tab:result2}: \abbr-base substantially outperforms all the other T5 variants and achieves comparable or even better results than the other base-size pre-trained models. Surprisingly, continually pre-training T5-base with SSR can achieve significant improvement over a transformer-big model pre-trained on rule-based synthetic data. We attribute this to the closer relationship between the task of GEC and our proposed SSR objective and more diverse grammatical errors introduced by the machine-generated spans. Interestingly, we observe the improvement of \abbr on the GEC task is even more significant than that on question generation and summarization datasets, because SSR is intuitively more similar to the challenge of GEC which can be well addressed by span correction \cite{chen2020improving}.

\subsection{Analysis}

\paragraph{Impact of Model Size} 
To analyze the effectiveness of the proposed SSR objective for Seq2Seq pre-trained models with different sizes, we report the performance comparison of small-size and large-size \abbr and different T5-based baselines. Note that we focus on analysis of SSR on the same T5 backbone model and do not compare against other large-sized Seq2Seq PTLMs because they are pre-trained with different data and number of stpes, thus are not comparable with our models. 

We present the results of large-size models and small-size models in Table \ref{tab:result3} and Table \ref{tab:result4}, respectively.\footnote{We do not compare against the variant with the denoising-based objective since its performance is consistently lower than the baseline in the previous experiments.} We find that the sequence span rewriting objective improves both large-size and small-size models. However, the improvement upon small-size models is significantly larger than that upon large-size models. This suggests that our method is more effective when the infilling model is significantly larger than the rewriting model. The performance of \abbr-small is also significantly better than DistilT5-small sequence-level knowledge distillation. That indicates SSR's potential on exploiting the knowledge from large pre-trained Seq2Seq transformers to improve the training of smaller models in a task-agnostic fashion.

\begin{table}[!tbp]
\centering
    \begin{minipage}{\linewidth}
	\centering
	\resizebox{0.95\linewidth}{!}{
		\begin{tabular}{llll}
			\toprule
\multirow{2}{*}{\textbf{Model}} & \multicolumn{3}{c}{\textbf{CNN/DM}} \\
			  & RG-1 & RG-2 & RG-L \\
			\midrule
			T5-small & 40.22 & 19.36 & 37.85 \\
			T5-small-cont & 40.43 & 19.55 & 38.08 \\
			DistilT5-small & 40.38 & 19.49 & 38.01 \\
            \abbr-small & \bf 41.95$^{*}$ & \bf 20.06$^{*}$ & \bf 39.01$^{*}$ \\
			\bottomrule
	\end{tabular}
	}
	\caption{Abstractive summarization results on CNN/DailyMail for SSR-small and corresponding T5 models of the same size.  $^{*}$The asterisk denotes statistically significant improvement with p-value < 0.05 upon all compared models.}
	\label{tab:result4}
	\end{minipage}
\end{table}

\begin{table}[!tbp]
\centering
    \begin{minipage}{\linewidth}
	\centering
	\resizebox{\linewidth}{!}{
		\begin{tabular}{llccc}
			\toprule
\multirow{2}{*}{\textbf{Model}} & \textbf{Imperfect} & \multicolumn{3}{c}{\textbf{CNN/DM}} \\
			  & \textbf{Span Generator} & RG-1 & RG-2 & RG-L \\
			\midrule
			T5-base & - & 42.25 & 20.22 & 39.45 \\
			SSR-base & T5-base & 42.78 & 20.51 & 39.97 \\
			SSR-base & T5-large & \bf 43.47 & \bf 20.74 & \bf 40.37 \\
			\midrule
			T5-small & - & 40.22 & 19.36 & 37.85 \\
			SSR-small & T5-base & 41.03 & 19.74 & 38.68 \\
			SSR-small & T5-large & \bf 41.95 & \bf 20.06 & \bf 39.01 \\
			\bottomrule
	\end{tabular}
	}
	\caption{Abstractive summarization results on CNN/DailyMail for SSR with imperfect span generator of different sizes.}
	\vspace{-3mm}
	\label{tab:result-gensize}
	\end{minipage}
\end{table}

\paragraph{Impact of Imperfect Span Generator} We also investigate the impact of the size of the imperfect span generator. This time, we generate imperfect text spans for pre-training using T5-base model and continually pre-train SSR-base and SSR-small. The results are shown in Table \ref{tab:result-gensize}. We find that our approach performs better with a larger imperfect span generator, which seems in contradiction to the findings in the replaced token detection objective introduced in ELECTRA~\cite{clark2020electra}. We suspect the reason is that the task of span-level infilling is more challenging than its token-level counterpart. Therefore, a small imperfect span generator may not be powerful enough to generate imperfect text spans that are meaningful and of relatively high quality. Consequently, the rewriting model may simply learn to ignore the imperfect spans and the SSR objective will degrade into text infilling. Moreover, we can see that the improvement yielded by the SSR objective is more significant when the size of the imperfect span generator is larger than the rewriting model that we aim to train. This confirms that SSR can effectively exploit the knowledge of a large model to better train a smaller one. Interestingly, we find that using imperfect spans generated by T5-base to continually pre-train T5-base can still improve the performance, which is similar to the case of self-distillation~\citep{furlanello2018born}.

\begin{table}[!tbp]
\centering
    \begin{minipage}{\linewidth}
	\centering
	\resizebox{\linewidth}{!}{
		\begin{tabular}{llll}
			\toprule
\multirow{2}{*}{\textbf{Model}} & \multicolumn{3}{c}{\textbf{CNN/DM}} \\
			  & RG-1 & RG-2 & RG-L \\
			\midrule
			SSR-base & \bf 43.53$^{*}$ & \bf 20.79$^{*}$ & \bf 40.47$^{*}$ \\
			\quad No curriculum &  43.26 & 20.53 & 40.14 \\	
			\quad Anti-curriculum & 43.09 & 20.48 & 40.01 \\
			\quad Loss-only curriculum & 43.40 & 20.67 & 40.25 \\
            \quad Length-only curriculum & 43.43 & 20.71 & 40.35 \\
			\bottomrule
	\end{tabular}
	}
	\caption{Ablation study results on CNN/DailyMail for SSR-base with different curriculum learning strategies. $^{*}$The asterisk denotes statistically significant improvement with p-value < 0.05 upon all compared ablation.}
	\label{tab:result-curr}
	\end{minipage}
\end{table}

\paragraph{Impact of Curriculum Pre-training} We then analyze the effectiveness of the proposed curriculum pre-training technique. We continually pre-train SSR-base with three variants of the proposed curriculum pre-training method: \textit{No curriculum} denotes the variant without curriculum pre-training; \textit{Anti-curriculum} denotes the variant where pre-training starts with difficult examples; \textit{Loss-only} and \textit{Length-only curriculum} denote a curriculum based solely on per-token loss and the length of imperfect span, respectively.
The results are shown in Table \ref{tab:result-curr}. We find that pre-training SSR from relatively easy examples to hard examples statistical significantly improve its performance on downstream tasks. More specifically, we find that scheduling the training examples by their length is slightly more effective than by per-token loss, while the combination of them can yield further improvements.

\section{Discussion}

\paragraph{Pre-training via Rewriting} We discuss several key advantages
of 
SSR
over the conventional text infilling objectives
here. \textbf{(1) SSR is closer to the downstream sequence transduction tasks.} This is because the model's prediction is not only based on its bidirectional context but also conditioned on the imperfect spans. In this way, the gap between pre-training and fine-tuning stages, which is introduced by the masked tokens or spans in conventional pre-training objectives, is alleviated. Indeed, many
NLG tasks
can be viewed as sequence span rewriting problems that rewrite the input text into another language, more compact format, grammatically correct sentences, or another style. 
\textbf{(2) SSR introduces more diverse noise patterns.} These patterns include paraphrasing and simplification of the text span, missing or redundant information, grammatical errors, and errors in terms of world knowledge or commonsense knowledge.
In fact, 
many of the rewriting patterns introduced by SSR resemble training examples in the downstream tasks. 
In contrast, conventional self-supervised Seq2Seq pre-training techniques rely 
on rule-based noise functions like text span masking, token masking, token deletion, token rotation, sentence shuffling, etc. \textbf{(3) SSR enables the model to learn from informative examples.} SSR enables the model to learn from informative examples, where the span generator makes an error. This provides more meaningful supervision and is also similar to the idea of active learning~\cite{settles2009active}. 

\paragraph{Distillation via Rewriting}
SSR sheds light on a new perspective of exploiting the knowledge of a large pre-trained model to improve smaller models. Similar to knowledge distillation (KD), this can be achieved by using a large-size teacher model pre-trained with the text infilling objective as the imperfect span generator, and pre-train or refine a small-size student model with the generated data using SSR. Different from conventional KD~\citep{hinton2015distilling} or sequence-level KD~\citep{kim2016sequence}, SSR enables the student model to exploit both teacher outputs and the ground truth at the same time. It is also related to boost learning~\citep{schapire2003boosting} and residual learning~\citep{he2016deep} in a sense that the model only needs to learn the prediction error of the teacher model, instead of the original task, text infilling, which may be too difficult for smaller-size models. 
\section{Conclusion}

We present sequence span rewriting (\abbr), a novel self-supervised objective for improving sequence-to-sequence transformers pre-trained with conventional text infilling objectives. SSR introduces more diverse and fine-grained learning signals and also bridges the gap between self-supervised pre-training and task-specific fine-tuning on common NLG datasets. Our experiments on continual T5 pre-training confirm the effectiveness of \abbr on improving pre-trained T5 models of different sizes across different tasks and datasets. 
Also, the large improvements achieved on small models with a larger imperfect span generator
indicates a new perspective of exploiting the knowledge of a large pre-trained model to help train smaller ones. 

\section*{Ethical Considerations}

Our approach is proposed to improve existing sequence-to-sequence pre-training techniques. It does not involve the collection and release of data except that generated by a pre-trained model, nor inference of information or judgments about individuals. That being said, since an improved sequence-to-sequence pre-trained model may be used in various downstream applications, it is still an important future direction to investigate the bias, fairness, and privacy issue in various kinds of pre-trained models. 
\section*{Acknowledgments}

We appreciate all anonymous reviewers and the meta-reviewer, for their insightful comments. Tao Ge is the corresponding author.
\bibliography{acl2020}

\begin{thebibliography}{62}
\expandafter\ifx\csname natexlab\endcsname\relax\def\natexlab#1{#1}\fi

\bibitem[{Banerjee and Lavie(2005)}]{banerjee-lavie-2005-meteor}
Satanjeev Banerjee and Alon Lavie. 2005.
\newblock \href {https://www.aclweb.org/anthology/W05-0909} {{METEOR}: An
  automatic metric for {MT} evaluation with improved correlation with human
  judgments}.
\newblock In \emph{Proceedings of the {ACL} Workshop on Intrinsic and Extrinsic
  Evaluation Measures for Machine Translation and/or Summarization}, pages
  65--72, Ann Arbor, Michigan. Association for Computational Linguistics.

\bibitem[{Bao et~al.(2020)Bao, Dong, Wei, Wang, Yang, Liu, Wang, Gao, Piao,
  Zhou, and Hon}]{bao2020unilmv2}
Hangbo Bao, Li~Dong, Furu Wei, Wenhui Wang, Nan Yang, Xiaodong Liu, Yu~Wang,
  Jianfeng Gao, Songhao Piao, Ming Zhou, and Hsiao{-}Wuen Hon. 2020.
\newblock \href {http://proceedings.mlr.press/v119/bao20a.html} {Unilmv2:
  Pseudo-masked language models for unified language model pre-training}.
\newblock In \emph{Proceedings of the 37th International Conference on Machine
  Learning, {ICML} 2020, 13-18 July 2020, Virtual Event}, volume 119 of
  \emph{Proceedings of Machine Learning Research}, pages 642--652. {PMLR}.

\bibitem[{Bender et~al.(2021)Bender, Gebru, McMillan{-}Major, and
  Shmitchell}]{parrot}
Emily~M. Bender, Timnit Gebru, Angelina McMillan{-}Major, and Shmargaret
  Shmitchell. 2021.
\newblock On the dangers of stochastic parrots: Can language models be too big?
\newblock In \emph{FAccT}, pages 610--623. {ACM}.

\bibitem[{Bengio et~al.(2009)Bengio, Louradour, Collobert, and
  Weston}]{bengio2009curriculum}
Yoshua Bengio, J{\'{e}}r{\^{o}}me Louradour, Ronan Collobert, and Jason Weston.
  2009.
\newblock \href {https://doi.org/10.1145/1553374.1553380} {Curriculum
  learning}.
\newblock In \emph{Proceedings of the 26th Annual International Conference on
  Machine Learning, {ICML} 2009, Montreal, Quebec, Canada, June 14-18, 2009},
  volume 382 of \emph{{ACM} International Conference Proceeding Series}, pages
  41--48. {ACM}.

\bibitem[{Bryant et~al.(2019)Bryant, Felice, Andersen, and
  Briscoe}]{bryant2019bea}
Christopher Bryant, Mariano Felice, {\O}istein~E. Andersen, and Ted Briscoe.
  2019.
\newblock \href {https://doi.org/10.18653/v1/W19-4406} {The {BEA}-2019 shared
  task on grammatical error correction}.
\newblock In \emph{Proceedings of the Fourteenth Workshop on Innovative Use of
  NLP for Building Educational Applications}, pages 52--75, Florence, Italy.
  Association for Computational Linguistics.

\bibitem[{Chen et~al.(2020)Chen, Ge, Zhang, Wei, and Zhou}]{chen2020improving}
Mengyun Chen, Tao Ge, Xingxing Zhang, Furu Wei, and Ming Zhou. 2020.
\newblock \href {https://doi.org/10.18653/v1/2020.emnlp-main.581} {Improving
  the efficiency of grammatical error correction with erroneous span detection
  and correction}.
\newblock In \emph{Proceedings of the 2020 Conference on Empirical Methods in
  Natural Language Processing (EMNLP)}, pages 7162--7169, Online. Association
  for Computational Linguistics.

\bibitem[{Clark et~al.(2020)Clark, Luong, Le, and Manning}]{clark2020electra}
Kevin Clark, Minh{-}Thang Luong, Quoc~V. Le, and Christopher~D. Manning. 2020.
\newblock \href {https://openreview.net/forum?id=r1xMH1BtvB} {{ELECTRA:}
  pre-training text encoders as discriminators rather than generators}.
\newblock In \emph{8th International Conference on Learning Representations,
  {ICLR} 2020, Addis Ababa, Ethiopia, April 26-30, 2020}. OpenReview.net.

\bibitem[{Dahlmeier et~al.(2013)Dahlmeier, Ng, and Wu}]{dahlmeier2013building}
Daniel Dahlmeier, Hwee~Tou Ng, and Siew~Mei Wu. 2013.
\newblock \href {https://www.aclweb.org/anthology/W13-1703} {Building a large
  annotated corpus of learner {E}nglish: The {NUS} corpus of learner
  {E}nglish}.
\newblock In \emph{Proceedings of the Eighth Workshop on Innovative Use of
  {NLP} for Building Educational Applications}, pages 22--31, Atlanta, Georgia.
  Association for Computational Linguistics.

\bibitem[{Devlin et~al.(2019)Devlin, Chang, Lee, and
  Toutanova}]{devlin2018bert}
Jacob Devlin, Ming-Wei Chang, Kenton Lee, and Kristina Toutanova. 2019.
\newblock \href {https://doi.org/10.18653/v1/N19-1423} {{BERT}: Pre-training of
  deep bidirectional transformers for language understanding}.
\newblock In \emph{Proceedings of the 2019 Conference of the North {A}merican
  Chapter of the Association for Computational Linguistics: Human Language
  Technologies, Volume 1 (Long and Short Papers)}, pages 4171--4186,
  Minneapolis, Minnesota. Association for Computational Linguistics.

\bibitem[{Dong et~al.(2019)Dong, Yang, Wang, Wei, Liu, Wang, Gao, Zhou, and
  Hon}]{dong2019unified}
Li~Dong, Nan Yang, Wenhui Wang, Furu Wei, Xiaodong Liu, Yu~Wang, Jianfeng Gao,
  Ming Zhou, and Hsiao{-}Wuen Hon. 2019.
\newblock \href
  {https://proceedings.neurips.cc/paper/2019/hash/c20bb2d9a50d5ac1f713f8b34d9aac5a-Abstract.html}
  {Unified language model pre-training for natural language understanding and
  generation}.
\newblock In \emph{Advances in Neural Information Processing Systems 32: Annual
  Conference on Neural Information Processing Systems 2019, NeurIPS 2019,
  December 8-14, 2019, Vancouver, BC, Canada}, pages 13042--13054.

\bibitem[{Du and Cardie(2018)}]{du-cardie-2018-harvesting}
Xinya Du and Claire Cardie. 2018.
\newblock \href {https://doi.org/10.18653/v1/P18-1177} {Harvesting
  paragraph-level question-answer pairs from {W}ikipedia}.
\newblock In \emph{Proceedings of the 56th Annual Meeting of the Association
  for Computational Linguistics (Volume 1: Long Papers)}, pages 1907--1917,
  Melbourne, Australia. Association for Computational Linguistics.

\bibitem[{Furlanello et~al.(2018)Furlanello, Lipton, Tschannen, Itti, and
  Anandkumar}]{furlanello2018born}
Tommaso Furlanello, Zachary~Chase Lipton, Michael Tschannen, Laurent Itti, and
  Anima Anandkumar. 2018.
\newblock \href {http://proceedings.mlr.press/v80/furlanello18a.html}
  {Born-again neural networks}.
\newblock In \emph{Proceedings of the 35th International Conference on Machine
  Learning, {ICML} 2018, Stockholmsm{\"{a}}ssan, Stockholm, Sweden, July 10-15,
  2018}, volume~80 of \emph{Proceedings of Machine Learning Research}, pages
  1602--1611. {PMLR}.

\bibitem[{Granger(1998)}]{granger1998computer}
Sylviane Granger. 1998.
\newblock \emph{The computer learner corpus: a versatile new source of data for
  SLA research}.
\newblock na.

\bibitem[{Grundkiewicz et~al.(2019)Grundkiewicz, Junczys-Dowmunt, and
  Heafield}]{grundkiewicz2019neural}
Roman Grundkiewicz, Marcin Junczys-Dowmunt, and Kenneth Heafield. 2019.
\newblock \href {https://doi.org/10.18653/v1/W19-4427} {Neural grammatical
  error correction systems with unsupervised pre-training on synthetic data}.
\newblock In \emph{Proceedings of the Fourteenth Workshop on Innovative Use of
  NLP for Building Educational Applications}, pages 252--263, Florence, Italy.
  Association for Computational Linguistics.

\bibitem[{He et~al.(2016)He, Zhang, Ren, and Sun}]{he2016deep}
Kaiming He, Xiangyu Zhang, Shaoqing Ren, and Jian Sun. 2016.
\newblock \href {https://doi.org/10.1109/CVPR.2016.90} {Deep residual learning
  for image recognition}.
\newblock In \emph{2016 {IEEE} Conference on Computer Vision and Pattern
  Recognition, {CVPR} 2016, Las Vegas, NV, USA, June 27-30, 2016}, pages
  770--778. {IEEE} Computer Society.

\bibitem[{Hermann et~al.(2015)Hermann, Kocisk{\'{y}}, Grefenstette, Espeholt,
  Kay, Suleyman, and Blunsom}]{hermann2015teaching}
Karl~Moritz Hermann, Tom{\'{a}}s Kocisk{\'{y}}, Edward Grefenstette, Lasse
  Espeholt, Will Kay, Mustafa Suleyman, and Phil Blunsom. 2015.
\newblock \href
  {https://proceedings.neurips.cc/paper/2015/hash/afdec7005cc9f14302cd0474fd0f3c96-Abstract.html}
  {Teaching machines to read and comprehend}.
\newblock In \emph{Advances in Neural Information Processing Systems 28: Annual
  Conference on Neural Information Processing Systems 2015, December 7-12,
  2015, Montreal, Quebec, Canada}, pages 1693--1701.

\bibitem[{Hinton et~al.(2015)Hinton, Vinyals, and Dean}]{hinton2015distilling}
Geoffrey Hinton, Oriol Vinyals, and Jeff Dean. 2015.
\newblock Distilling the knowledge in a neural network.
\newblock \emph{arXiv preprint arXiv:1503.02531}.

\bibitem[{Holtzman et~al.(2020)Holtzman, Buys, Du, Forbes, and
  Choi}]{Holtzman2020The}
Ari Holtzman, Jan Buys, Li~Du, Maxwell Forbes, and Yejin Choi. 2020.
\newblock \href {https://openreview.net/forum?id=rygGQyrFvH} {The curious case
  of neural text degeneration}.
\newblock In \emph{8th International Conference on Learning Representations,
  {ICLR} 2020, Addis Ababa, Ethiopia, April 26-30, 2020}. OpenReview.net.

\bibitem[{Hu et~al.(2017)Hu, Yang, Liang, Salakhutdinov, and
  Xing}]{DBLP:conf/icml/HuYLSX17}
Zhiting Hu, Zichao Yang, Xiaodan Liang, Ruslan Salakhutdinov, and Eric~P. Xing.
  2017.
\newblock \href {http://proceedings.mlr.press/v70/hu17e.html} {Toward
  controlled generation of text}.
\newblock In \emph{Proceedings of the 34th International Conference on Machine
  Learning, {ICML} 2017, Sydney, NSW, Australia, 6-11 August 2017}, volume~70
  of \emph{Proceedings of Machine Learning Research}, pages 1587--1596. {PMLR}.

\bibitem[{Joshi et~al.(2020)Joshi, Chen, Liu, Weld, Zettlemoyer, and
  Levy}]{DBLP:journals/tacl/JoshiCLWZL20}
Mandar Joshi, Danqi Chen, Yinhan Liu, Daniel~S. Weld, Luke Zettlemoyer, and
  Omer Levy. 2020.
\newblock \href {https://doi.org/10.1162/tacl_a_00300} {{S}pan{BERT}: Improving
  pre-training by representing and predicting spans}.
\newblock \emph{Transactions of the Association for Computational Linguistics},
  8:64--77.

\bibitem[{Kim and Rush(2016)}]{kim2016sequence}
Yoon Kim and Alexander~M. Rush. 2016.
\newblock \href {https://doi.org/10.18653/v1/D16-1139} {Sequence-level
  knowledge distillation}.
\newblock In \emph{Proceedings of the 2016 Conference on Empirical Methods in
  Natural Language Processing}, pages 1317--1327, Austin, Texas. Association
  for Computational Linguistics.

\bibitem[{Kiyono et~al.(2019)Kiyono, Suzuki, Mita, Mizumoto, and
  Inui}]{kiyono2019empirical}
Shun Kiyono, Jun Suzuki, Masato Mita, Tomoya Mizumoto, and Kentaro Inui. 2019.
\newblock \href {https://doi.org/10.18653/v1/D19-1119} {An empirical study of
  incorporating pseudo data into grammatical error correction}.
\newblock In \emph{Proceedings of the 2019 Conference on Empirical Methods in
  Natural Language Processing and the 9th International Joint Conference on
  Natural Language Processing (EMNLP-IJCNLP)}, pages 1236--1242, Hong Kong,
  China. Association for Computational Linguistics.

\bibitem[{Lan et~al.(2020)Lan, Chen, Goodman, Gimpel, Sharma, and
  Soricut}]{lan2019albert}
Zhenzhong Lan, Mingda Chen, Sebastian Goodman, Kevin Gimpel, Piyush Sharma, and
  Radu Soricut. 2020.
\newblock \href {https://openreview.net/forum?id=H1eA7AEtvS} {{ALBERT:} {A}
  lite {BERT} for self-supervised learning of language representations}.
\newblock In \emph{8th International Conference on Learning Representations,
  {ICLR} 2020, Addis Ababa, Ethiopia, April 26-30, 2020}. OpenReview.net.

\bibitem[{Lewis et~al.(2020{\natexlab{a}})Lewis, Ghazvininejad, Ghosh,
  Aghajanyan, Wang, and Zettlemoyer}]{lewis2020pre}
Mike Lewis, Marjan Ghazvininejad, Gargi Ghosh, Armen Aghajanyan, Sida Wang, and
  Luke Zettlemoyer. 2020{\natexlab{a}}.
\newblock \href
  {https://proceedings.neurips.cc/paper/2020/hash/d6f1dd034aabde7657e6680444ceff62-Abstract.html}
  {Pre-training via paraphrasing}.
\newblock In \emph{Advances in Neural Information Processing Systems 33: Annual
  Conference on Neural Information Processing Systems 2020, NeurIPS 2020,
  December 6-12, 2020, virtual}.

\bibitem[{Lewis et~al.(2020{\natexlab{b}})Lewis, Liu, Goyal, Ghazvininejad,
  Mohamed, Levy, Stoyanov, and Zettlemoyer}]{lewis2019bart}
Mike Lewis, Yinhan Liu, Naman Goyal, Marjan Ghazvininejad, Abdelrahman Mohamed,
  Omer Levy, Veselin Stoyanov, and Luke Zettlemoyer. 2020{\natexlab{b}}.
\newblock \href {https://doi.org/10.18653/v1/2020.acl-main.703} {{BART}:
  Denoising sequence-to-sequence pre-training for natural language generation,
  translation, and comprehension}.
\newblock In \emph{Proceedings of the 58th Annual Meeting of the Association
  for Computational Linguistics}, pages 7871--7880, Online. Association for
  Computational Linguistics.

\bibitem[{Lin et~al.(2020)Lin, Zhou, Shen, Zhou, Bhagavatula, Choi, and
  Ren}]{DBLP:conf/emnlp/LinZSZBCR20}
Bill~Yuchen Lin, Wangchunshu Zhou, Ming Shen, Pei Zhou, Chandra Bhagavatula,
  Yejin Choi, and Xiang Ren. 2020.
\newblock \href {https://doi.org/10.18653/v1/2020.findings-emnlp.165}
  {Commongen: {A} constrained text generation challenge for generative
  commonsense reasoning}.
\newblock In \emph{Proceedings of the 2020 Conference on Empirical Methods in
  Natural Language Processing: Findings, {EMNLP} 2020, Online Event, 16-20
  November 2020}, pages 1823--1840. Association for Computational Linguistics.

\bibitem[{Lin(2004)}]{lin-2004-rouge}
Chin-Yew Lin. 2004.
\newblock \href {https://www.aclweb.org/anthology/W04-1013} {{ROUGE}: A package
  for automatic evaluation of summaries}.
\newblock In \emph{Text Summarization Branches Out}, pages 74--81, Barcelona,
  Spain. Association for Computational Linguistics.

\bibitem[{Liu and Lapata(2019)}]{liu2019text}
Yang Liu and Mirella Lapata. 2019.
\newblock \href {https://doi.org/10.18653/v1/D19-1387} {Text summarization with
  pretrained encoders}.
\newblock In \emph{Proceedings of the 2019 Conference on Empirical Methods in
  Natural Language Processing and the 9th International Joint Conference on
  Natural Language Processing (EMNLP-IJCNLP)}, pages 3730--3740, Hong Kong,
  China. Association for Computational Linguistics.

\bibitem[{Liu et~al.(2019)Liu, Ott, Goyal, Du, Joshi, Chen, Levy, Lewis,
  Zettlemoyer, and Stoyanov}]{liu2019roberta}
Yinhan Liu, Myle Ott, Naman Goyal, Jingfei Du, Mandar Joshi, Danqi Chen, Omer
  Levy, Mike Lewis, Luke Zettlemoyer, and Veselin Stoyanov. 2019.
\newblock Roberta: A robustly optimized bert pretraining approach.
\newblock \emph{arXiv preprint arXiv:1907.11692}.

\bibitem[{Michel et~al.(2019)Michel, Levy, and Neubig}]{michel2019sixteen}
Paul Michel, Omer Levy, and Graham Neubig. 2019.
\newblock \href
  {https://proceedings.neurips.cc/paper/2019/hash/2c601ad9d2ff9bc8b282670cdd54f69f-Abstract.html}
  {Are sixteen heads really better than one?}
\newblock In \emph{Advances in Neural Information Processing Systems 32: Annual
  Conference on Neural Information Processing Systems 2019, NeurIPS 2019,
  December 8-14, 2019, Vancouver, BC, Canada}, pages 14014--14024.

\bibitem[{Mizumoto et~al.(2011)Mizumoto, Komachi, Nagata, and
  Matsumoto}]{mizumoto2011mining}
Tomoya Mizumoto, Mamoru Komachi, Masaaki Nagata, and Yuji Matsumoto. 2011.
\newblock \href {https://www.aclweb.org/anthology/I11-1017} {Mining revision
  log of language learning {SNS} for automated {J}apanese error correction of
  second language learners}.
\newblock In \emph{Proceedings of 5th International Joint Conference on Natural
  Language Processing}, pages 147--155, Chiang Mai, Thailand. Asian Federation
  of Natural Language Processing.

\bibitem[{Narayan et~al.(2018)Narayan, Cohen, and Lapata}]{narayan2018don}
Shashi Narayan, Shay~B. Cohen, and Mirella Lapata. 2018.
\newblock \href {https://doi.org/10.18653/v1/D18-1206} {Don{'}t give me the
  details, just the summary! topic-aware convolutional neural networks for
  extreme summarization}.
\newblock In \emph{Proceedings of the 2018 Conference on Empirical Methods in
  Natural Language Processing}, pages 1797--1807, Brussels, Belgium.
  Association for Computational Linguistics.

\bibitem[{Ng et~al.(2014)Ng, Wu, Briscoe, Hadiwinoto, Susanto, and
  Bryant}]{ng2014conll}
Hwee~Tou Ng, Siew~Mei Wu, Ted Briscoe, Christian Hadiwinoto, Raymond~Hendy
  Susanto, and Christopher Bryant. 2014.
\newblock \href {https://doi.org/10.3115/v1/W14-1701} {The {C}o{NLL}-2014
  shared task on grammatical error correction}.
\newblock In \emph{Proceedings of the Eighteenth Conference on Computational
  Natural Language Learning: Shared Task}, pages 1--14, Baltimore, Maryland.
  Association for Computational Linguistics.

\bibitem[{Papineni et~al.(2002)Papineni, Roukos, Ward, and
  Zhu}]{papineni-etal-2002-bleu}
Kishore Papineni, Salim Roukos, Todd Ward, and Wei-Jing Zhu. 2002.
\newblock \href {https://doi.org/10.3115/1073083.1073135} {{B}leu: a method for
  automatic evaluation of machine translation}.
\newblock In \emph{Proceedings of the 40th Annual Meeting of the Association
  for Computational Linguistics}, pages 311--318, Philadelphia, Pennsylvania,
  USA. Association for Computational Linguistics.

\bibitem[{Peters et~al.(2018)Peters, Neumann, Iyyer, Gardner, Clark, Lee, and
  Zettlemoyer}]{peters2018deep}
Matthew Peters, Mark Neumann, Mohit Iyyer, Matt Gardner, Christopher Clark,
  Kenton Lee, and Luke Zettlemoyer. 2018.
\newblock \href {https://doi.org/10.18653/v1/N18-1202} {Deep contextualized
  word representations}.
\newblock In \emph{Proceedings of the 2018 Conference of the North {A}merican
  Chapter of the Association for Computational Linguistics: Human Language
  Technologies, Volume 1 (Long Papers)}, pages 2227--2237, New Orleans,
  Louisiana. Association for Computational Linguistics.

\bibitem[{Radford et~al.(2018)Radford, Narasimhan, Salimans, and
  Sutskever}]{radford2018improving}
Alec Radford, Karthik Narasimhan, Tim Salimans, and Ilya Sutskever. 2018.
\newblock Improving language understanding by generative pre-training.

\bibitem[{Raffel et~al.(2019)Raffel, Shazeer, Roberts, Lee, Narang, Matena,
  Zhou, Li, and Liu}]{raffel2019exploring}
Colin Raffel, Noam Shazeer, Adam Roberts, Katherine Lee, Sharan Narang, Michael
  Matena, Yanqi Zhou, Wei Li, and Peter~J Liu. 2019.
\newblock Exploring the limits of transfer learning with a unified text-to-text
  transformer.
\newblock \emph{arXiv preprint arXiv:1910.10683}.

\bibitem[{Rajpurkar et~al.(2016)Rajpurkar, Zhang, Lopyrev, and
  Liang}]{rajpurkar2016squad}
Pranav Rajpurkar, Jian Zhang, Konstantin Lopyrev, and Percy Liang. 2016.
\newblock \href {https://doi.org/10.18653/v1/D16-1264} {{SQ}u{AD}: 100,000+
  questions for machine comprehension of text}.
\newblock In \emph{Proceedings of the 2016 Conference on Empirical Methods in
  Natural Language Processing}, pages 2383--2392, Austin, Texas. Association
  for Computational Linguistics.

\bibitem[{Romero et~al.(2015)Romero, Ballas, Kahou, Chassang, Gatta, and
  Bengio}]{romero2014fitnets}
Adriana Romero, Nicolas Ballas, Samira~Ebrahimi Kahou, Antoine Chassang, Carlo
  Gatta, and Yoshua Bengio. 2015.
\newblock \href {http://arxiv.org/abs/1412.6550} {Fitnets: Hints for thin deep
  nets}.
\newblock In \emph{3rd International Conference on Learning Representations,
  {ICLR} 2015, San Diego, CA, USA, May 7-9, 2015, Conference Track
  Proceedings}.

\bibitem[{Sanh et~al.(2019)Sanh, Debut, Chaumond, and
  Wolf}]{sanh2019distilbert}
Victor Sanh, Lysandre Debut, Julien Chaumond, and Thomas Wolf. 2019.
\newblock Distilbert, a distilled version of bert: smaller, faster, cheaper and
  lighter.
\newblock \emph{arXiv preprint arXiv:1910.01108}.

\bibitem[{Schapire(2003)}]{schapire2003boosting}
Robert~E Schapire. 2003.
\newblock The boosting approach to machine learning: An overview.
\newblock In \emph{Nonlinear estimation and classification}, pages 149--171.
  Springer.

\bibitem[{Schwartz et~al.(2020)Schwartz, Stanovsky, Swayamdipta, Dodge, and
  Smith}]{schwartz2020right}
Roy Schwartz, Gabriel Stanovsky, Swabha Swayamdipta, Jesse Dodge, and Noah~A.
  Smith. 2020.
\newblock \href {https://doi.org/10.18653/v1/2020.acl-main.593} {The right tool
  for the job: Matching model and instance complexities}.
\newblock In \emph{Proceedings of the 58th Annual Meeting of the Association
  for Computational Linguistics}, pages 6640--6651, Online. Association for
  Computational Linguistics.

\bibitem[{See et~al.(2017)See, Liu, and Manning}]{see2017get}
Abigail See, Peter~J. Liu, and Christopher~D. Manning. 2017.
\newblock \href {https://doi.org/10.18653/v1/P17-1099} {Get to the point:
  Summarization with pointer-generator networks}.
\newblock In \emph{Proceedings of the 55th Annual Meeting of the Association
  for Computational Linguistics (Volume 1: Long Papers)}, pages 1073--1083,
  Vancouver, Canada. Association for Computational Linguistics.

\bibitem[{Settles(2009)}]{settles2009active}
Burr Settles. 2009.
\newblock Active learning literature survey.

\bibitem[{Shen et~al.(2020)Shen, Dong, Ye, Ma, Yao, Gholami, Mahoney, and
  Keutzer}]{shen2020q}
Sheng Shen, Zhen Dong, Jiayu Ye, Linjian Ma, Zhewei Yao, Amir Gholami,
  Michael~W Mahoney, and Kurt Keutzer. 2020.
\newblock Q-bert: Hessian based ultra low precision quantization of bert.
\newblock In \emph{AAAI}, pages 8815--8821.

\bibitem[{Song et~al.(2019)Song, Tan, Qin, Lu, and Liu}]{song2019mass}
Kaitao Song, Xu~Tan, Tao Qin, Jianfeng Lu, and Tie{-}Yan Liu. 2019.
\newblock \href {http://proceedings.mlr.press/v97/song19d.html} {{MASS:} masked
  sequence to sequence pre-training for language generation}.
\newblock In \emph{Proceedings of the 36th International Conference on Machine
  Learning, {ICML} 2019, 9-15 June 2019, Long Beach, California, {USA}},
  volume~97 of \emph{Proceedings of Machine Learning Research}, pages
  5926--5936. {PMLR}.

\bibitem[{Strubell et~al.(2019)Strubell, Ganesh, and
  McCallum}]{strubell2019energy}
Emma Strubell, Ananya Ganesh, and Andrew McCallum. 2019.
\newblock \href {https://doi.org/10.18653/v1/P19-1355} {Energy and policy
  considerations for deep learning in {NLP}}.
\newblock In \emph{Proceedings of the 57th Annual Meeting of the Association
  for Computational Linguistics}, pages 3645--3650, Florence, Italy.
  Association for Computational Linguistics.

\bibitem[{Vaswani et~al.(2017)Vaswani, Shazeer, Parmar, Uszkoreit, Jones,
  Gomez, Kaiser, and Polosukhin}]{vaswani2017attention}
Ashish Vaswani, Noam Shazeer, Niki Parmar, Jakob Uszkoreit, Llion Jones,
  Aidan~N. Gomez, Lukasz Kaiser, and Illia Polosukhin. 2017.
\newblock \href
  {https://proceedings.neurips.cc/paper/2017/hash/3f5ee243547dee91fbd053c1c4a845aa-Abstract.html}
  {Attention is all you need}.
\newblock In \emph{Advances in Neural Information Processing Systems 30: Annual
  Conference on Neural Information Processing Systems 2017, December 4-9, 2017,
  Long Beach, CA, {USA}}, pages 5998--6008.

\bibitem[{Vedantam et~al.(2015)Vedantam, Zitnick, and
  Parikh}]{vedantam2015cider}
Ramakrishna Vedantam, C.~Lawrence Zitnick, and Devi Parikh. 2015.
\newblock \href {https://doi.org/10.1109/CVPR.2015.7299087} {Cider:
  Consensus-based image description evaluation}.
\newblock In \emph{{IEEE} Conference on Computer Vision and Pattern
  Recognition, {CVPR} 2015, Boston, MA, USA, June 7-12, 2015}, pages
  4566--4575. {IEEE} Computer Society.

\bibitem[{Wang et~al.(2019)Wang, Zhao, Jia, Li, and
  Liu}]{wang-etal-2019-denoising}
Liang Wang, Wei Zhao, Ruoyu Jia, Sujian Li, and Jingming Liu. 2019.
\newblock \href {https://doi.org/10.18653/v1/D19-1412} {Denoising based
  sequence-to-sequence pre-training for text generation}.
\newblock In \emph{Proceedings of the 2019 Conference on Empirical Methods in
  Natural Language Processing and the 9th International Joint Conference on
  Natural Language Processing (EMNLP-IJCNLP)}, pages 4003--4015, Hong Kong,
  China. Association for Computational Linguistics.

\bibitem[{Wolf et~al.(2020)Wolf, Debut, Sanh, Chaumond, Delangue, Moi, Cistac,
  Rault, Louf, Funtowicz, Davison, Shleifer, von Platen, Ma, Jernite, Plu, Xu,
  Le~Scao, Gugger, Drame, Lhoest, and Rush}]{wolf2020transformers}
Thomas Wolf, Lysandre Debut, Victor Sanh, Julien Chaumond, Clement Delangue,
  Anthony Moi, Pierric Cistac, Tim Rault, Remi Louf, Morgan Funtowicz, Joe
  Davison, Sam Shleifer, Patrick von Platen, Clara Ma, Yacine Jernite, Julien
  Plu, Canwen Xu, Teven Le~Scao, Sylvain Gugger, Mariama Drame, Quentin Lhoest,
  and Alexander Rush. 2020.
\newblock \href {https://doi.org/10.18653/v1/2020.emnlp-demos.6} {Transformers:
  State-of-the-art natural language processing}.
\newblock In \emph{Proceedings of the 2020 Conference on Empirical Methods in
  Natural Language Processing: System Demonstrations}, pages 38--45, Online.
  Association for Computational Linguistics.

\bibitem[{Xu et~al.(2020)Xu, Zhou, Ge, Wei, and Zhou}]{xu2020bert}
Canwen Xu, Wangchunshu Zhou, Tao Ge, Furu Wei, and Ming Zhou. 2020.
\newblock \href {https://doi.org/10.18653/v1/2020.emnlp-main.633}
  {{BERT}-of-theseus: Compressing {BERT} by progressive module replacing}.
\newblock In \emph{Proceedings of the 2020 Conference on Empirical Methods in
  Natural Language Processing (EMNLP)}, pages 7859--7869, Online. Association
  for Computational Linguistics.

\bibitem[{Yang et~al.(2019)Yang, Dai, Yang, Carbonell, Salakhutdinov, and
  Le}]{yang2019xlnet}
Zhilin Yang, Zihang Dai, Yiming Yang, Jaime~G. Carbonell, Ruslan Salakhutdinov,
  and Quoc~V. Le. 2019.
\newblock \href
  {https://proceedings.neurips.cc/paper/2019/hash/dc6a7e655d7e5840e66733e9ee67cc69-Abstract.html}
  {Xlnet: Generalized autoregressive pretraining for language understanding}.
\newblock In \emph{Advances in Neural Information Processing Systems 32: Annual
  Conference on Neural Information Processing Systems 2019, NeurIPS 2019,
  December 8-14, 2019, Vancouver, BC, Canada}, pages 5754--5764.

\bibitem[{Yannakoudakis et~al.(2011)Yannakoudakis, Briscoe, and
  Medlock}]{yannakoudakis2011new}
Helen Yannakoudakis, Ted Briscoe, and Ben Medlock. 2011.
\newblock \href {https://www.aclweb.org/anthology/P11-1019} {A new dataset and
  method for automatically grading {ESOL} texts}.
\newblock In \emph{Proceedings of the 49th Annual Meeting of the Association
  for Computational Linguistics: Human Language Technologies}, pages 180--189,
  Portland, Oregon, USA. Association for Computational Linguistics.

\bibitem[{Zellers et~al.(2019)Zellers, Holtzman, Rashkin, Bisk, Farhadi,
  Roesner, and Choi}]{zellers2019defending}
Rowan Zellers, Ari Holtzman, Hannah Rashkin, Yonatan Bisk, Ali Farhadi,
  Franziska Roesner, and Yejin Choi. 2019.
\newblock \href
  {https://proceedings.neurips.cc/paper/2019/hash/3e9f0fc9b2f89e043bc6233994dfcf76-Abstract.html}
  {Defending against neural fake news}.
\newblock In \emph{Advances in Neural Information Processing Systems 32: Annual
  Conference on Neural Information Processing Systems 2019, NeurIPS 2019,
  December 8-14, 2019, Vancouver, BC, Canada}, pages 9051--9062.

\bibitem[{Zhang and Bansal(2019)}]{zhang2019addressing}
Shiyue Zhang and Mohit Bansal. 2019.
\newblock \href {https://doi.org/10.18653/v1/D19-1253} {Addressing semantic
  drift in question generation for semi-supervised question answering}.
\newblock In \emph{Proceedings of the 2019 Conference on Empirical Methods in
  Natural Language Processing and the 9th International Joint Conference on
  Natural Language Processing (EMNLP-IJCNLP)}, pages 2495--2509, Hong Kong,
  China. Association for Computational Linguistics.

\bibitem[{Zhou et~al.(2020{\natexlab{a}})Zhou, Ge, Mu, Xu, Wei, and
  Zhou}]{zhou2019improving}
Wangchunshu Zhou, Tao Ge, Chang Mu, Ke~Xu, Furu Wei, and Ming Zhou.
  2020{\natexlab{a}}.
\newblock \href {https://doi.org/10.18653/v1/2020.findings-emnlp.30} {Improving
  grammatical error correction with machine translation pairs}.
\newblock In \emph{Findings of the Association for Computational Linguistics:
  EMNLP 2020}, pages 318--328, Online. Association for Computational
  Linguistics.

\bibitem[{Zhou et~al.(2020{\natexlab{b}})Zhou, Ge, and
  Xu}]{zhou-etal-2020-pseudo}
Wangchunshu Zhou, Tao Ge, and Ke~Xu. 2020{\natexlab{b}}.
\newblock \href {https://doi.org/10.18653/v1/2020.findings-emnlp.136}
  {Pseudo-bidirectional decoding for local sequence transduction}.
\newblock In \emph{Findings of the Association for Computational Linguistics:
  EMNLP 2020}, pages 1506--1511, Online. Association for Computational
  Linguistics.

\bibitem[{Zhou et~al.(2021{\natexlab{a}})Zhou, Lee, Selvam, Lee, and
  Ren}]{DBLP:conf/iclr/ZhouLSL021}
Wangchunshu Zhou, Dong{-}Ho Lee, Ravi~Kiran Selvam, Seyeon Lee, and Xiang Ren.
  2021{\natexlab{a}}.
\newblock \href {https://openreview.net/forum?id=3k20LAiHYL2} {Pre-training
  text-to-text transformers for concept-centric common sense}.
\newblock In \emph{9th International Conference on Learning Representations,
  {ICLR} 2021, Virtual Event, Austria, May 3-7, 2021}. OpenReview.net.

\bibitem[{Zhou et~al.(2020{\natexlab{c}})Zhou, Xu, Ge, McAuley, Xu, and
  Wei}]{zhou2020bert}
Wangchunshu Zhou, Canwen Xu, Tao Ge, Julian McAuley, Ke~Xu, and Furu Wei.
  2020{\natexlab{c}}.
\newblock Bert loses patience: Fast and robust inference with early exit.
\newblock \emph{In NeurIPS 2020}.

\bibitem[{Zhou et~al.(2021{\natexlab{b}})Zhou, Xu, and
  McAuley}]{DBLP:journals/corr/abs-2106-04570}
Wangchunshu Zhou, Canwen Xu, and Julian~J. McAuley. 2021{\natexlab{b}}.
\newblock \href {http://arxiv.org/abs/2106.04570} {Meta learning for knowledge
  distillation}.
\newblock \emph{CoRR}, abs/2106.04570.

\bibitem[{Zhu et~al.(2015)Zhu, Kiros, Zemel, Salakhutdinov, Urtasun, Torralba,
  and Fidler}]{zhu2015aligning}
Yukun Zhu, Ryan Kiros, Richard~S. Zemel, Ruslan Salakhutdinov, Raquel Urtasun,
  Antonio Torralba, and Sanja Fidler. 2015.
\newblock \href {https://doi.org/10.1109/ICCV.2015.11} {Aligning books and
  movies: Towards story-like visual explanations by watching movies and reading
  books}.
\newblock In \emph{2015 {IEEE} International Conference on Computer Vision,
  {ICCV} 2015, Santiago, Chile, December 7-13, 2015}, pages 19--27. {IEEE}
  Computer Society.

\end{thebibliography}
\bibliographystyle{acl_natbib}

\end{document}